\documentclass[runningheads]{llncs}

\usepackage{accv}

\usepackage{accvabbrv}

\usepackage{graphicx}
\usepackage{booktabs}
\usepackage{tabularx}
\usepackage{epsfig}
\usepackage{graphicx}
\usepackage{amsmath}
\usepackage{amssymb}
\usepackage{colortbl}
\usepackage{multirow}
\usepackage{color}
\usepackage{setspace}
\usepackage[normalem]{ulem}
\usepackage{tikz}
\usepackage{comment}
\usepackage{algorithm}
\usepackage{algpseudocode}
\usepackage{float}
\usepackage{caption}
\usepackage{adjustbox}
\usepackage{pifont}%
\usepackage{wrapfig}

\definecolor{top1}{RGB}{245,152,153}
\definecolor{top2}{RGB}{253,205,154}
\definecolor{top3}{RGB}{248,244,140}

\usepackage[accsupp]{axessibility}  %

\definecolor{blgray}{gray}{0.97}
\definecolor{mygray}{gray}{.93}

\usepackage{amsmath}

\usepackage[accsupp]{axessibility}  %

\usepackage[pagebackref,breaklinks,colorlinks,citecolor=accvblue]{hyperref}

\usepackage[capitalize]{cleveref}
\crefname{section}{Sec.}{Secs.}
\Crefname{section}{Section}{Sections}
\Crefname{table}{Table}{Tables}
\crefname{table}{Tab.}{Tabs.}

\usepackage{orcidlink}

\makeatletter
\def\hlinew#1{%
  \noalign{\ifnum0=`}\fi\hrule \@height #1 \futurelet
   \reserved@a\@xhline}
\makeatother

\begin{document}

\title{High-Quality Visually-Guided Sound Separation from Diverse Categories} 

\titlerunning{DAVIS: {D}iffsuion-based Audio-VIsual Separation}

\author{Chao Huang\inst{1}\and
Susan Liang\inst{1}\and
Yapeng Tian\inst{1}\and \\
Anurag Kumar\inst{2} \and
Chenliang Xu\inst{1}
}

\authorrunning{C. Huang et al.}

\institute{University of Rochester, Rochester NY 14627, USA \and
Meta Reality Labs Research, Redmond WA 98052, USA\\}

\maketitle

\begin{abstract}
We propose DAVIS, a \textbf{D}iffusion-based \textbf{A}udio-\textbf{VI}sual \textbf{S}epara-tion framework that solves the audio-visual sound source separation task through generative learning. 
Existing methods typically frame sound separation as a mask-based regression problem, achieving significant progress. 
However, they face limitations in capturing the complex data distribution required for high-quality separation of sounds from diverse categories. In contrast, DAVIS leverages a generative diffusion model and a Separation U-Net to synthesize separated sounds directly from Gaussian noise, conditioned on both the audio mixture and the visual information. With its generative objective, DAVIS is better suited to achieving the goal of high-quality sound separation across diverse sound categories. We compare DAVIS to existing state-of-the-art discriminative audio-visual separation methods on the AVE and MUSIC datasets, and results show that DAVIS outperforms other methods in separation quality, demonstrating the advantages of our framework for tackling the audio-visual source separation task. 
Our project page is available here: \url{https://wikichao.github.io/data/projects/DAVIS/}.

\end{abstract}

\section{Introduction}
Visually-guided sound source separation, also referred to as audio-visual separation, is a pivotal task for assessing a machine perception system's ability to understand multisensory signals~\cite{zhao2018sound,gao2018learning}. 
It aims to separate individual sounds from a complex audio mixture by utilizing visual cues about the objects that are producing the sounds, \textit{e.g.}, separate the ``barking'' sound from the mixture by querying the ``dog'' object. An effective separation model should be capable of handling a \textbf{\textit{diverse}} range of sounds and producing \textbf{\textit{high-quality}} separations that can deliver a realistic auditory experience.
The community has devoted considerable effort to tackling this task~\cite{zhao2018sound,gao2019co,gan2020music,chatterjee2021visual,tian2021cyclic,dong2023clipsep,zhu2022visually,chen2023iquery}, developing more powerful separation frameworks~\cite{zhao2018sound,gao2019co,chatterjee2021visual,chen2023iquery}, proposing more effective training pipelines~\cite{tian2021cyclic}, and incorporating additional visual cues~\cite{gan2020music} to enhance the performance. Conventional approaches usually employ discriminative learning through mask regression~\cite{zhao2018sound} or spectrogram reconstruction~\cite{owens2018audio} as training objectives.

\begin{figure*}[!tbp]
    \centering
    \includegraphics[width=0.9\textwidth]{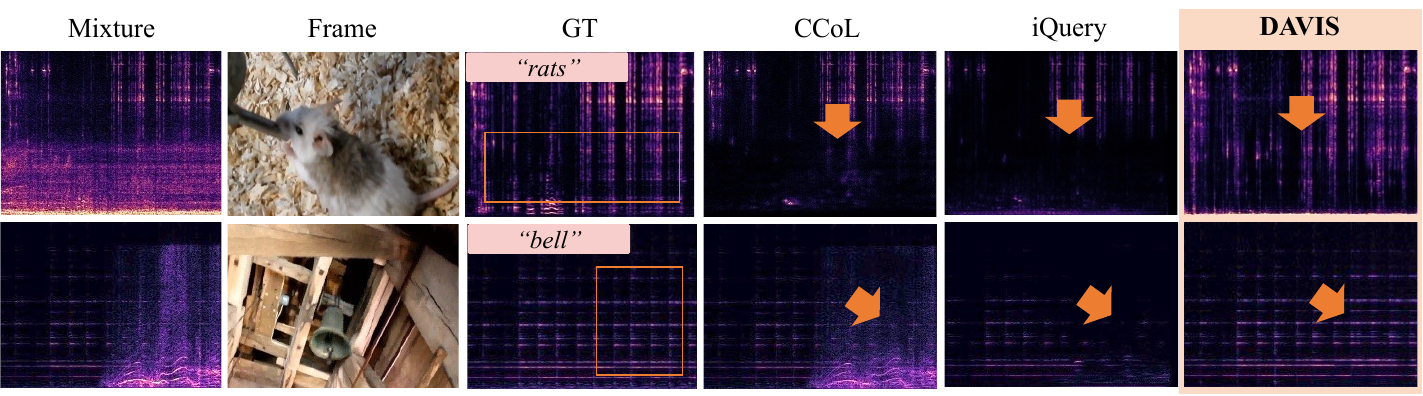}
    \caption{
    \textbf{Separation results on diverse time-frequency structures} are shown for SOTA discriminative methods and our proposed DAVIS. Each row displays the audio mixture, reference visual frame, ground truth magnitude, and predicted magnitudes from DAVIS, iQuery~\cite{chen2023iquery}, and CCoL~\cite{tian2021cyclic}. DAVIS successfully recovers suppressed time-frequency structures (highlighted in the box), where mask-regression methods fail.
    }
    \label{fig:motivating_examples}
\end{figure*}

While these methods have shown promising separation performance, they are inherently limited in dealing with diverse time-frequency structures and separating sounds in challenging situations. For instance, different sounds can interact in complicated ways, and sometimes the desired sound is heavily suppressed by the others (\eg, examples shown in \cref{fig:motivating_examples}), posing significant hurdles in regressing the mask from hidden sound patterns.
Therefore, a natural question arises: \textit{is there an effective approach to model complex data distributions, capture precise audio-visual associations, and generate high-quality separated sounds?}

We answer the question by introducing a generative framework for audio-visual separation. A new class of generative models called denoising diffusion probabilistic models (DDPMs)~\cite{ho2020denoising,nichol2021improved,song2020denoising}, also known as diffusion models, has emerged recently and demonstrated remarkable abilities in generating diverse and high-quality images~\cite{dhariwal2021diffusion} and audio~\cite{kong2020diffwave}.
Nonetheless, at first glance, whether diffusion models can be effectively repurposed for audio-visual separation remains unclear. The challenges of this task stem from the necessity of specialized architecture to capture audio-visual correspondences coupled with modeling the unique characteristics of magnitude spectrograms. 
Generic diffusion models may not be well-suited to this task without addressing the above challenges.

In this paper, we present DAVIS, a novel diffusion model-based audio-visual separation framework.
Unlike conventional methods that regress masks, DAVIS tackles separation as a conditional generation process, iteratively ``growing'' the magnitude spectrogram from Gaussian noise to the desired sound. This approach distributes the burden of recovering complex time-frequency patterns across multiple steps, making DAVIS versatile for serving various scenarios, even the challenging cases shown in~\cref{fig:motivating_examples}.
A key component to train the diffusion model for audio-visual separation is the network architecture. Given the complexities of spectrograms and audio-visual correlations, we propose a Separation U-Net, well-suited for capturing these features.
In particular, we find that modeling long-range dependencies is important as similar but distant time frames commonly exist. 
We, therefore, introduce a Convolution-Attention (CA) block in the Separation U-Net to capture both local and non-local contexts.
Furthermore, to enhance the audio-visual association learning, we explore different interaction manners and devise a Feature Interaction module to facilitate the injection of visual cues into the separation task.

Another challenge arises from the frequent occurrence of silent time frames in magnitude spectrograms, where the values are almost zero. This skewed data distribution renders the conventional $\mathcal{L}_2$ loss in diffusion models susceptible to error. However, these silent parts also provide valuable information during inference, indicating the overlap between the mixed and target sounds. Therefore, they can be effectively utilized in the sampling process. To address these issues, we propose using a more robust $\mathcal{L}_1$ loss for training and a silence mask-guided sampling strategy to refine the results at the inference stage.

Experiments on the AVE~\cite{tian2018audio} and MUSIC~\cite{zhao2018sound} datasets demonstrate that DAVIS consistently outperforms the state-of-the-art methods in terms of separation quality. Our contributions are summarized as follows:
\begin{itemize}
    \item We approach the audio-visual separation task as a conditional generation process with generative diffusion models. %
    \item  We highlight the importance of network design and propose a Separation U-Net, equipping with an Audio-Visual Feature Interaction Module to capture multimodal association effectively.  
    \item We identify an issue in magnitude spectrograms that can be leveraged to enhance the inference process and propose a novel silence mask-guided sampling strategy as a solution.
    \item Our framework is competitive with or surpasses previous methods on datasets with diverse categories, validating the effectiveness of our approach.
\end{itemize}

\section{Related Work}
\noindent\textbf{Audio-Visual Sound Source Separation.}
In this section, our focus is on modern audio-visual sound source separation approaches while acknowledging the prolonged research efforts dedicated to signal processing-based separation~\cite{virtanen2007monaural,smaragdis2003non} and other multimodal research~\cite{NEURIPS2023_760dff0f, liang2023neural, chen2024RAF}.
Recent deep learning-based audio-visual sound source separation methods have been applied to a variety of audio categories, including speech signals~\cite{ephrat2018looking,owens2018audio,afouras2020self,michelsanti2021overview}, musical instrument sounds~\cite{zhao2018sound,gan2020music,tian2021cyclic,gao2019co,zhao2019sound,chatterjee2021visual,tan2023language}, and universal sound sources~\cite{gao2018learning,mittal2022learning,tzinis2020into,tzinis2022audioscopev2,chatterjee2022learning,zhu2022visually,dong2023clipsep,chen2023iquery}. These methods typically employ a learning regime that involves mixing two audio streams from different videos to provide supervised training signals. A sound separation network, often implemented as a U-Net, is then used for mask regression~\cite{zhao2018sound} conditioned on the associated visual features. In recent years, research in this area has focused on both domain-specific and open-domain sound source separation~\cite{tzinis2020into,mittal2022learning,zhu2022visually,dong2023clipsep,chen2023iquery}. However, existing methods often require additional information, such as text queries~\cite{dong2023clipsep}, motion cues~\cite{mittal2022learning,zhu2022visually}, or class labels~\cite{chen2023iquery}, to achieve satisfactory performance. In this paper, we propose a novel generative audio-visual separation approach that demonstrates competitive or outperforms existing methods in separating both specific and open-domain sound sources.

\noindent\textbf{Diffusion Models.}
Diffusion models~\cite{ho2020denoising,song2020score,song2019generative} fall under the category of deep generative models that start with a sample in a random distribution and gradually restore the data sample through a denoising process. Recently, diffusion models have exhibited remarkable performance across various domains, including computer vision~\cite{dhariwal2021diffusion,avrahami2022blended,ramesh2022hierarchical,gu2022vector,nichol2021glide,ho2022video,singer2022make,ruiz2022dreambooth,saharia2022photorealistic}, natural language processing~\cite{austin2021structured,gong2022diffuseq,li2022diffusion,chen2022analog}, audio applications~\cite{kong2020diffwave,popov2021grad,lee2021nu,chen2022analog,chen2020wavegrad,huang2022prodiff,scheibler2023diffusion}, and audio-visual content generation~\cite{ruan2022mm}.
Furthermore, there has been a growing interest in utilizing diffusion models for discriminative tasks. Some pioneer works have explored the application of diffusion models to image segmentation~\cite{amit2021segdiff,baranchuk2021label,brempong2022denoising} and object detection~\cite{chen2022diffusiondet}. Despite the significant interest in this direction, successful applications of generative diffusion models to audio-visual scene understanding remain limited. A few recent works have attempted to use diffusion-based approaches for audio-visual speech separation and enhancement~\cite{lee2023seeing,chou2023av2wav}. 
However, these works limit themselves to speech separation or enhancement and do not study the more challenging audio-visual sound separation problem. The major blockers are the inherent challenges in designing effective network architectures to capture audio-visual correspondence and adapting diffusion models to handle unique data distributions.
In this paper, we address this gap by employing a diffusion model for audio-visual sound separation. Our novel separation architecture empowers the diffusion-based model to effectively learn the intricate relationships between audio and visual modalities, leading to superior separation performance.

\begin{figure*}[t]
\begin{center}
\includegraphics[width=0.8\linewidth]{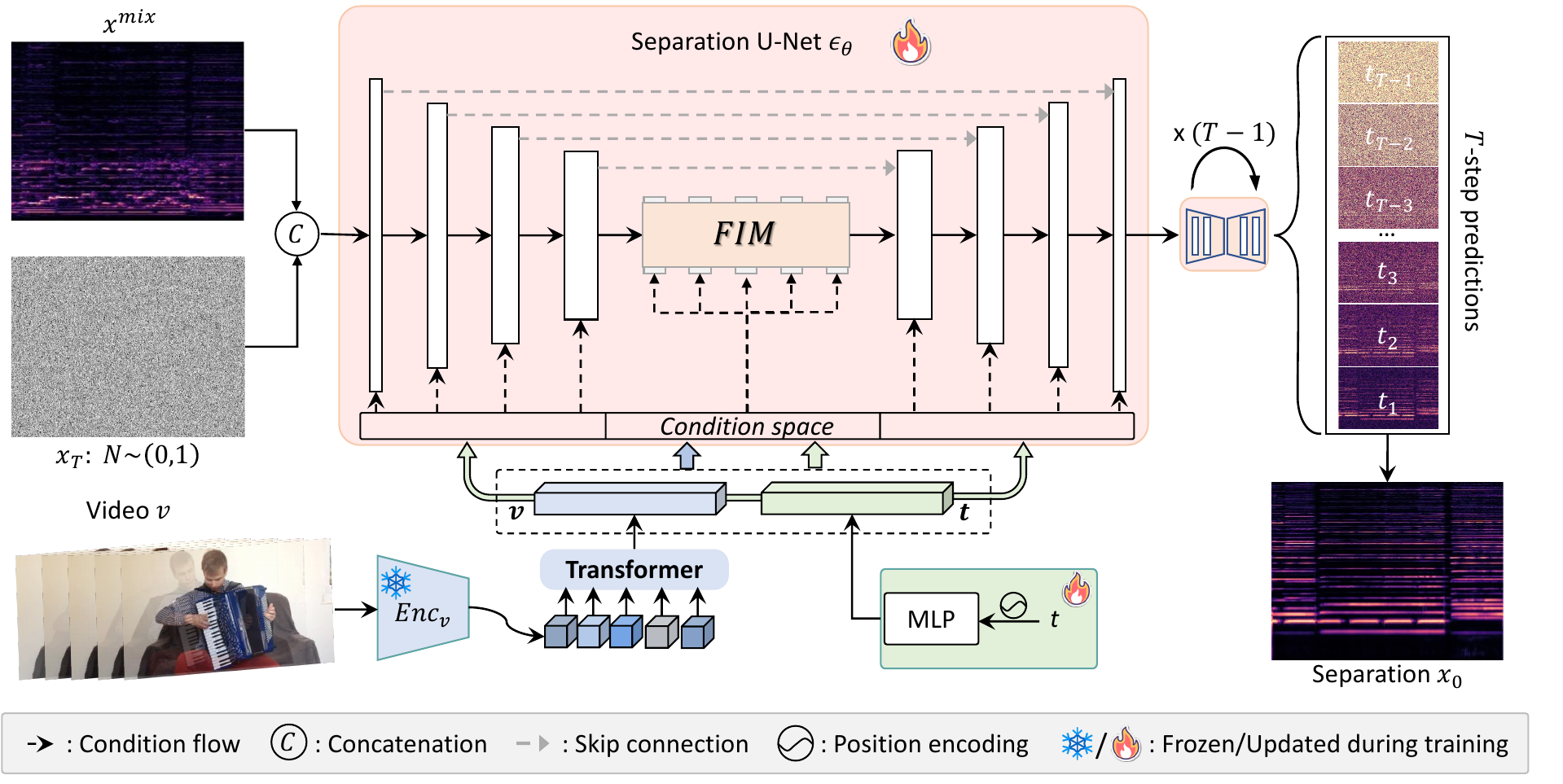}
\end{center}
  \caption{\textbf{Overview of the DAVIS framework.} 
We aim to synthesize $x_0$ from the mixture $x^{mix}$, visual stream $v$, and timestep $t$. Starting with $x_T$ from a standard distribution, we encode $v=\{I_j\}_{j=1}^K$ and $t$ into the embedding space. A temporal transformer generates the visual feature $\boldsymbol{v}$, which, along with $\boldsymbol{t}$, conditions the Separation U-Net $\epsilon\theta$ to iteratively denoise $x_T$ into $x_0$. $\boldsymbol{v}$ is used only in the Feature Interaction Module for audio-visual association, while $\boldsymbol{t}$ is used throughout.
  }
\label{fig: framework}
\end{figure*}

\section{Method}
In this section, we introduce DAVIS, our novel diffusion model-based audio-visual separation framework. We begin by providing a brief recap of diffusion models in \cref{subsec:preliminaries}. Then, we describe our task setup and give a method overview in \cref{subsec:setup}. Next, we present our proposed separation framework in \cref{subsec:network}. Furthermore, we discuss the training pipeline in \cref{subsec:training}. Finally, we introduce the silence mask-guided sampling strategy in \cref{subsec:sampling}.

\subsection{Preliminaries: Diffusion Models}
\label{subsec:preliminaries}
Diffusion models~\cite{ho2020denoising} typically consist of a forward and a reverse process. The forward process is defined as a Markov chain that gradually adds noise to the data sample $x_0$ according to a variance schedule $\beta_1, ..., \beta_T$. To sample $x_t$ at an arbitrary timestep $t$, we have:
\begin{equation}
    q(x_t | x_0) = \mathcal{N}(x_t; \sqrt{\Bar{\alpha}_t}x_{0}, (1-\Bar{\alpha}_t)\mathbf{I}),
\label{eq:1}
\end{equation}
where $\alpha_t = 1 - \beta_t$ and $\Bar{\alpha}_t = \prod_{s=0}^t \alpha_s$. Note that the variance schedule is fixed for both forward and reverse processes. If the total number of $T$ goes to infinity, the diffusion process will finally lead to pure noise, \textit{i.e.}, the distribution of $p(x_T) $ is $\mathcal{N}(x_t;\mathbf{0}, \mathbf{I})$ with only Gaussian noise.

The reverse process aims to recover samples from Gaussian distribution by removing the noise gradually, which is a Markov chain parameterized by $\theta$:
\begin{equation}
    p_\theta(x_{0:T}) = p(x_{T})\prod_{t=1}^T p_\theta(x_{t-1}|x_t),
\label{eq:2}
\end{equation}
where at each iteration, the transition is formulated as:

\begin{equation}
    p_\theta(x_{t-1}|x_t) = \mathcal{N}(x_{t-1};\boldsymbol{\mu_\theta}(x_t, t, \boldsymbol{c}), \Tilde{\beta_t}\mathbf{I}).
\label{eq:3}    
\end{equation}
Note that we set the variances to untrained constants $\Tilde{\beta_t}\mathbf{I}$ following~\cite{ho2020denoising}, and $\boldsymbol{\mu_\theta}(x_t, t, \boldsymbol{c})$ is typically implemented as neural networks. Unlike vanilla diffusion models, 
we include the conditional context $\boldsymbol{c}$ as additional network inputs, which represent audio mixture and visual information in our task.

To train the network, a simplified way is to penalize the $\epsilon$-prediction with the $\mathcal{L}_2$ loss, which is equal to predict $\boldsymbol{\mu_\theta}$ according to its parameterization~\cite{ho2020denoising}:
\begin{equation}
    \mathcal{L}_2(\theta) = \mathbb{E}_{t, x_0, \epsilon}[ ||\epsilon - \epsilon_\theta(\sqrt{\Bar{\alpha}_t}x_0 +\sqrt{1-\Bar{\alpha}_t}\epsilon, \boldsymbol{c}, t )||^2].
\label{eq:4}    
\end{equation}
Here, $\epsilon_\theta$ represents a function approximator used to predict the noise added at each iteration, while $t$ is a uniformly sampled value ranging from 1 to $T$.

\subsection{Task Setup and Method Overview}
\label{subsec:setup}
Given an unlabeled video clip $V$, we can extract an audio-visual pair $(a, v)$, where $a$ and $v$ are the audio and visual streams, respectively. In real-world scenarios, the audio stream can be a mixture of $N$ individual sound sources, denoted as $a = \sum_{i=1}^N s_i$, where each source $s_i$ can be of various categories. Meanwhile, the visual stream $v$ is typically a synchronized video of $K$ frames, denoted as $v=\{I_j\}_{j=1}^K$. The visually-guided sound source separation task aims to utilize visual cues from $v$ to help separate $a$ into $N$ individual sources $s_i$. Since no labels are provided to distinguish the sound sources $s_i$, prior works~\cite{zhao2018sound,tian2021cyclic,huang2023egocentric} have commonly used a ``mix and separate'' strategy, which involves mixing audio streams from two different videos and manually create the mixture: $a^{mix} = a^{(1)} + a^{(2)}$. In practice, audio is usually transformed into magnitude spectrogram by short-time Fourier transform $x = \mathbf{STFT}(a) \in \mathbb{R}^{ T \times F}$, allowing for manipulations in the 2D Time-Frequency domain. Here, $F$ and $T$ are the numbers of frequency bins and time frames, respectively. Consequently, the goal of training is to learn a separation network capable of mapping $\boldsymbol{f: (x^{mix}, v^{(i)}) \rightarrow x^{(i)}}$. For simplicity, we will omit the video index notation in the subsequent sections\footnote{In this paper, superscripts denote video indices, while subscripts usually refer to diffusion timesteps.}.

In contrast to conventional approaches that perform the mapping through regression, our proposed DAVIS framework is built on a diffusion model with a $T$-step diffusion and reverse processes. The diffusion process is determined by a fixed variance schedule as described in \cref{eq:1}, which gradually adds noises to the magnitude spectrogram $x_0$ and converts it to latent $x_T$.
As depicted in \cref{fig: framework}, the reverse process (according to \cref{eq:2} and \cref{eq:3}) of DAVIS is specified by our proposed Separation U-Net $\epsilon_\theta$. 
This reverse process iteratively denoises a latent variable $x_T$, which is sampled from a uniform distribution, to obtain a separated magnitude conditioned on the magnitude of the input sound mixture $x^{mix}$ and the visual stream $v$. Consequently, the objective of the Separation U-Net $\epsilon_\theta$ is to predict the noise $\epsilon$ added at each diffusion timestep during forward.

\subsection{Proposed DAVIS Framework }
\label{subsec:network}

Diffusion models often use U-Net-like~\cite{ronneberger2015u} architectures, which excel at capturing multi-level feature representations and maintaining the output shape identical to the input. These properties also make them well-suited for the audio-visual separation task in the network aspect. However, naively applying existing conditional diffusion models to audio-visual separation is ineffective, as they are typically designed for image-to-image translation~\cite{meng2021sdedit} or text-to-image synthesis~\cite{saharia2022photorealistic,rombach2022high}. These models utilize different condition mechanisms than those required for audio-visual tasks, and they are not tailored to address the unique characteristics of audio-visual data. Therefore, the development of a specialized audio-visual separation network for diffusion models is essential. 
In this context, we revisit the challenges that need to be addressed: (1) Similar frequency patterns commonly exist even in temporally distant time frames, which necessitates the network to capture both long-range dependencies across time and frequency dimensions, and thus pure convolution~\cite{zhao2018sound,gao2019co} may fall short. (2) Real-world videos often have mismatched visual and audio content. Extracting visual condition~\cite{tian2021cyclic,dong2023clipsep} without considering the possible unrelated audio-visual content can potentially lead to less discriminative visual cues. (3) Establishing precise audio-visual associations is crucial, but directly concatenating visual and audio embeddings at the bottleneck~\cite{gao2019co} lacks the ability to foster further interactions between the two modalities. 

To address these challenges, we propose a novel and specialized Separation U-Net in our diffusion model that incorporates Convolution-Attention blocks to learn both local and global time-frequency associations, introduce a simple yet effective temporal transformer to aggregate the frame features and devise an audio-visual feature interaction module to enhance association learning by enabling interactions between audio and visual modalities.

\begin{figure}[t]
\begin{center}
\includegraphics[width=1\linewidth]{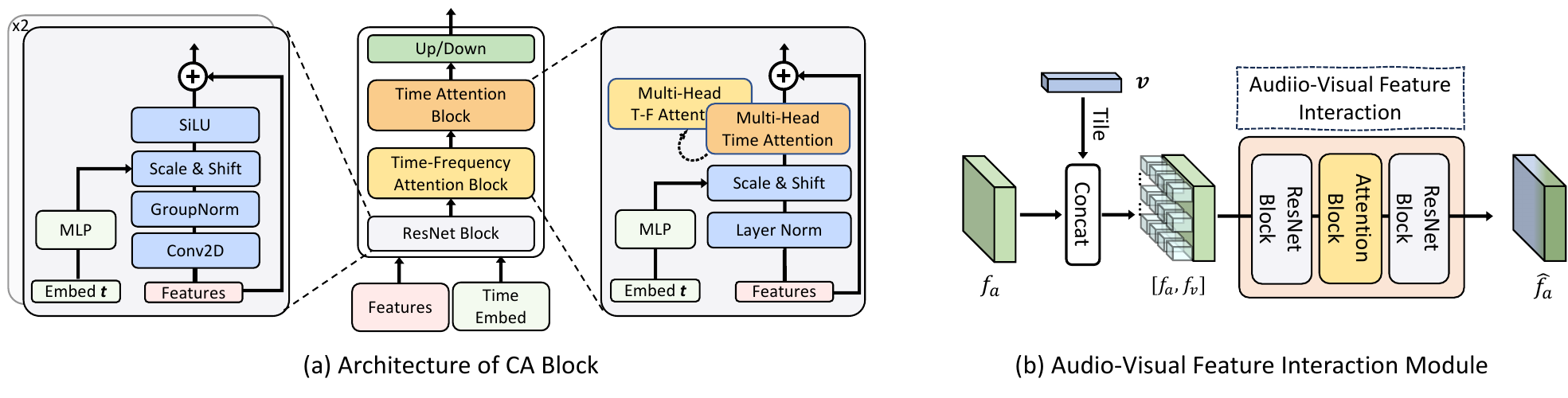}
\end{center}
  \caption{\textbf{Illustrations on (a) CA block: } 
  It operates by taking audio feature maps and a time embedding $\boldsymbol{t}$ as inputs. Each sub-block, except the up/down sampling layer, is conditioned on $\boldsymbol{t}$. 
    ResNet and attention blocks are stacked to capture local and non-local audio contexts;
    \textbf{(b) Audio-Visual Feature Interaction Module:} It functions by replicating and concatenating $\boldsymbol{v}$ with $\boldsymbol{f_a}$, and uses two identical ResNet blocks and an attention block to process the concatenated features.
    }
\label{fig: ca}
\end{figure}

\noindent\textbf{Encoder/Decoder Designs.} Our proposed Separation U-Net architecture consists of an encoder and a decoder, linked by an audio-visual feature interaction module. Both the encoder and decoder comprise five CA blocks. Initially, we concatenate the latent variable $x_T$ with the mixture $x^{mix}$ along the channel dimension and use a 1x1 convolution to project it into the feature space (another 1x1 convolution to convert the decoder output back to magnitude). As depicted in \cref{fig: ca}(a), each CA block consists of a ResNet block, a Time-Frequency Attention block, and a Time Attention block.
Following this, a down-sample or an up-sample layer with a scale factor of 2 is used. 
Concretely, we build the ResNet block using WeightStandardized 2D convolution~\cite{qian2020multiple} along with GroupNormalization~\cite{wu2018group} and SiLU activation~\cite{elfwing2018sigmoid}. To incorporate the time embedding $\boldsymbol{t}$ as a condition, a Multi-Layer Perceptron (MLP) is used to generate $\boldsymbol{t}$-dependent scaling and shifting vectors for feature-wise affine transformation~\cite{dumoulin2018feature}. 
We also adopt an efficient form of attention mechanism~\cite{shen2021efficient} for implementing the Time-Frequency Attention block. 
To enhance the long-range time dependency modeling, a Time Attention block is then appended. In practice, we follow the design in \cite{wang2023tf}, which includes Pre-Layer Normalization and Multi-Head Attention along the time dimension within the residual connection. 
The down-sample and up-sample layers are simply 2D convolutions with a stride of 2. As a result, we can obtain audio feature maps $\boldsymbol{f_a} \in \mathbb{R}^{C \times \frac{T}{32} \times \frac{F}{32}}$ at the bottleneck, where $C$ represents the number of channels. 

\noindent\textbf{Timestep Embedding.} In a diffusion model, the timestep embedding serves to inform the model about the current position within the Markov chain. As shown in \cref{fig: framework}, timestep $t$ is specified by the sinusoidal positional encoding~\cite{vaswani2017attention} and further transformed by an MLP, which will be passed to each CA block as a timestep condition.

\noindent\textbf{Visual Condition Aggregation.}
Not all frames in a video will be attributable to the synchronized audio. To account for unaligned visual content, we incorporate a shallow transformer to effectively aggregate the visual condition. Concretely, we extract frame features $\{\boldsymbol{I}_j\}_{j=1}^K$
from the visual stream $v$ using a pre-trained visual backbone $\mathbf{Enc_v}$, where $\boldsymbol{I}_j \in \mathbb{R}^{C}$. 
We apply a self-attention temporal transformer $\phi(\cdot)$ to aggregate raw visual frame features, resulting in $\{\boldsymbol{\hat{I}}_j\}_{j=1}^K = \phi(\{\boldsymbol{I}_j\}_{j=1}^K)$. For the transformer design, we empirically find that a shallow transformer with three encoder layers and one decoder layer works well. The global visual embedding $\boldsymbol{v}$ is then computed by averaging the temporal dimension of $\boldsymbol{v} = \frac{1}{K}\sum_{j=1}^K{\boldsymbol{\hat{I}}_j}$.

\noindent\textbf{Audio-Visual Feature Interaction Module.} The key to audio-visual separation lies in effectively utilizing visual information to separate visually-indicated sound sources. Therefore, the interaction between audio and visual modalities at the feature level becomes crucial. Existing approaches often concatenate audio and visual features at the bottleneck~\cite{gao2019co,chatterjee2021visual} and pass them to the decoder for further fusion. This design, however, imposes a dual task on the decoder: to integrate visual cues while simultaneously reconstructing the audio signal. We hypothesize that enabling further audio-visual interaction at the bottleneck could potentially enhance the separation performance. To this end, we explore different interaction manners and propose an audio-visual feature interaction module to improve this capability (see \cref{tab:fim}). 
We spatially tile $\boldsymbol{v}$ to match the shape of $\boldsymbol{f_a}$, resulting in visual feature maps $\boldsymbol{f_v}$.  Subsequently, the audio and visual feature maps are concatenated along channel dimension and fed into the feature interaction module (FIM): $\hat{\boldsymbol{f_a}}:= \mathbf{FIM}([\boldsymbol{f_a}, \boldsymbol{f_v}])$,
where $\hat{\boldsymbol{f_a}} \in \mathbb{R}^{C \times \frac{T}{32} \times \frac{F}{32}}$. The details of the module are illustrated in \cref{fig: ca}(b), including two ResNet blocks and a Time-Frequency Attention block to facilitate capturing audio-visual associations within both local and global regions.

\subsection{Training Pipeline}
\label{subsec:training}
Given the sampled audio-visual pairs from the dataset, we first adopt the ``mix and separate'' strategy and compute the magnitudes $x^{(1)}, x^{(2)}, x^{mix}$ with STFT. To align with the frequency decomposition of the human auditory system, we apply a logarithmic transformation to the magnitude spectrogram, converting it to a log-frequency scale. Additionally, we ensure consistent scaling by multiplying log-frequency magnitudes with a scale factor $\sigma$ and clipping the values to fall within the range [0, 1].

\noindent\textbf{$\mathcal{L}_1$ Denosing Loss.}
The visual frames are encoded to embeddings $\boldsymbol{v^{(1)}}$,$\boldsymbol{v^{(2)}}$. Taking video (1) as an example, we sample $\epsilon$ from a standard Gaussian distribution and $t$ from the set $\{1,..., T\}$. Then, we input $x^{(1)}_t, x^{mix},\boldsymbol{v}^{(1)}, t$ to the Separation U-Net $\epsilon_\theta$ and optimize the network by taking a gradient step on \cref{eq:4}. 
While $\mathcal{L}_2$ loss is well-suited for Gaussian noise estimation, the distribution of magnitude spectrograms is usually left-skewed due to the silent time frames and various frequency patterns, resulting in numerous regions with near-zero values. We hypothesize that the $\mathcal{L}_1$ loss is more robust to this type of data distribution and, therefore, be beneficial for training the denoising neural network.
In practice, we use both videos (1) and (2) for training, and the final loss term is formulated as 
$\mathcal{L} =\mathcal{L}^{(1)}_{1}(\theta) +\mathcal{L}^{(2)}_{1}(\theta)$.

\subsection{Silence Mask-Guided Sampling}
\label{subsec:sampling}
Our inference process starts from a sampled latent variable $x_T$, and ends with a sample from the target distribution $p(x_0|x^{mix},\boldsymbol{v})$, conditioned on the mixture and visual frame embedding. As the goal of separation is to 
predict the individual sound from the mixture, an observation is drawn: silent time frames in the mixture should also be silent in the separated sound, \textit{i.e.}, the network can leverage portions of the mixture $x^{mix}$ to sample an individual sound $x_0$.

Given a sampling step from time $t$ to $t-1$, the transition distribution can be rewrited using \cref{eq:1} on the $x^{mix}$ and \cref{eq:3} on the $x_t$:
\begin{subequations}
    \begin{align}
        x_{t-1}^{{mix}} &\sim \mathcal{N}(\sqrt{\bar{\alpha}_{t-1}}x^{mix}, (1 - \bar{\alpha}_{t-1})\mathbf{I}), \label{eq:sub1}\\
        x_{t-1} &\sim \mathcal{N}(\boldsymbol{\mu_\theta}(x_t, t, \boldsymbol{c}), \Tilde{\beta_t}\mathbf{I}), \label{eq:sub2}\\
        m &= [x^{mix} < \delta_{silence}], \label{eq:sub3}\\
        {\Hat{x}_{t-1}} &= m \odot x_{t-1}^{mix} + (1 - m) \odot x_{t-1}. \label{eq:sub4}
    \end{align}
\end{subequations}
The refined output ${\Hat{x}_{t-1}}$ is fed into the subsequent sampling iteration, further reducing the distribution gap between the ground truth target sound and the prediction. We introduce the hyperparameter $\delta_{silence}$ as the threshold for determining the silence region, setting it to a fixed value of 0.002. With the above modification, the model is enforced to predict individual sounds as parts of the mixture akin to the conventional mask-based approaches. Thus, the model refrains from "hallucinating" content as generation tasks, and the separation performance is improved as well (see \cref{tab:silence}).
The waveform $s_i$ for the separated sound can be reconstructed by applying inverse STFT on the magnitude prediction and the original mixture phase.

\begin{table}[!t]
    \centering
    \footnotesize
    \begin{tabularx}{0.8\textwidth}{lXrrrXrrr}
        \toprule
        & & \multicolumn{3}{c}{AVE~\cite{tian2018audio}}                                        & & \multicolumn{3}{c}{MUSIC~\cite{zhao2018sound}} \\
                            \cmidrule(lr){3-5}                                                        \cmidrule(lr){7-9}
        Methods    & & SDR $\uparrow$    & SIR $\uparrow$    & SAR $\uparrow$ & & SDR $\uparrow$    & SIR $\uparrow$    & SAR $\uparrow$\\
        \midrule
        NMF-MFCC${}^\dag$~\cite{spiertz2009source} & &- & -& -& & $0.92$ & $5.68$ & $6.84$ \\
        Sound-of-Pixels${}^\dag$~\cite{zhao2018sound} & & $1.21$ & $7.08$ & $6.84$ & & $4.23$ & $9.39$ & $9.85$ \\
        Co-Separation${}^\dag$~\cite{gao2019co} & &- & -& -& & $6.54$ & $11.37$ & $9.46$ \\
        Sound-of-Motions${}^\dag$~\cite{zhao2019sound} & & $1.48$ & $7.41$ & $7.39$  & &- & -& - \\
        Minus-Plus${}^\dag$~\cite{xu2019recursive} & & $1.96$ & $7.95$ & $8.08$ & &- & -& - \\
        Cascaded Filter${}^\dag$~\cite{zhu2020visually} & & $2.68$ & $8.18$ & $8.48$ & &- & -& - \\
        CCoL${}^\dag$~\cite{tian2021cyclic} & &- & -& -& & \cellcolor{top3}$7.74$ & \cellcolor{top3}$13.22$ & \cellcolor{top3}$11.54$ \\
        AMnet${}^\dag$~\cite{zhu2022visually} & & \cellcolor{top3}$3.71$ & \cellcolor{top1}$9.15$ & \cellcolor{top2}$11.00$ & &- & -& -\\
        iQuery${}^\dag$~\cite{chen2023iquery} & & \cellcolor{top1}$5.02$ & \cellcolor{top3}$8.21$ & \cellcolor{top1}$12.32$ & & \cellcolor{top2}$11.17$ & \cellcolor{top2}$15.84$ & \cellcolor{top2}$14.27$ \\
        \midrule
         DAVIS (ours) & & \cellcolor{top2}$4.86$  & \cellcolor{top2}$9.13$  & \cellcolor{top3}$9.92$ & & \cellcolor{top1}$\mathbf{11.61}$ & \cellcolor{top1}$\mathbf{18.36}$ & \cellcolor{top1}$\mathbf{14.70}$ \\
        \bottomrule
    \end{tabularx}
    \caption{Comparison of our method to other audio-visual separation approaches on the AVE and MUSIC test set. The top three results
    are highlighted in \textcolor{top1}{red}, \textcolor{top2}{orange}, and \textcolor{top3}{yellow}, respectively. The
    results noted by ${}^\dag$ are obtained from \cite{chen2023iquery,zhu2022visually}. 
Note that audio in AVE could include off-screen sounds and background noise, which may reduce the accuracy of the reported metrics.}
    \label{tab:ave_music}
\end{table}

\begin{figure*}[t]
    \centering
    \includegraphics[width=0.8\textwidth]{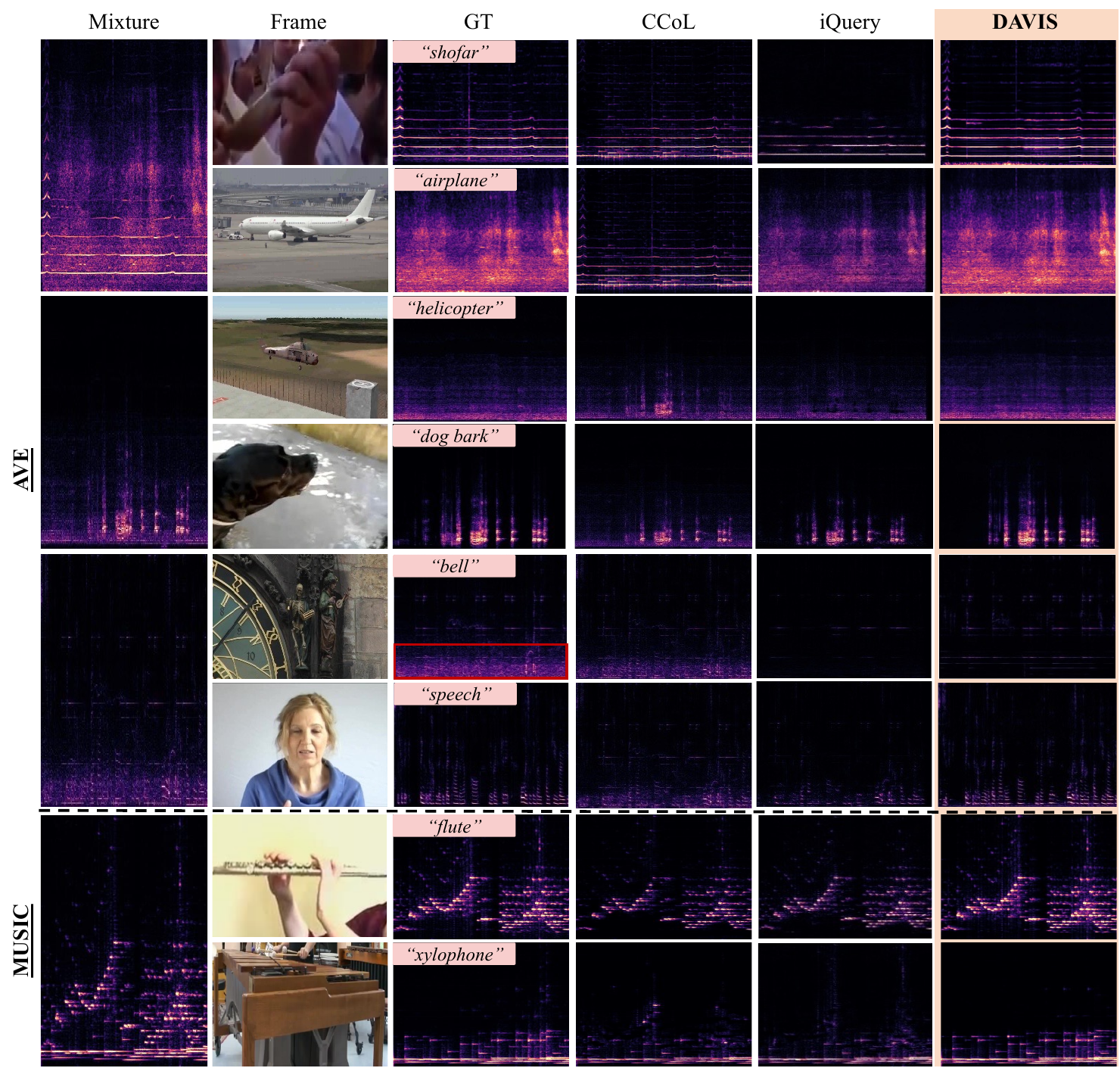}
    \caption{Visualizations of audio-visual separation results on the AVE (the top three mixtures) and MUSIC (the last mixture) datasets. Two sounds are mixed, and reference frames are provided to guide the separation. The comparison is shown between the predictions made by DAVIS (ours), iQuery~\cite{chen2023iquery}, and CCoL~\cite{tian2021cyclic} with the ground truth. DAVIS can effectively separate sound mixtures from various categories, such as \textit{airplane}, \textit{rats}, and \textit{dog barking}.}
    \label{fig:vis}
\end{figure*}

\section{Experiments}

\subsection{Experimental Setup}

\noindent \textbf{Datasets.}
Our model demonstrates the ability to handle mixtures of diverse sound categories. To evaluate our approach, we use  AVE~\cite{tian2018audio} and MUSIC~\cite{zhao2018sound} datasets, which cover musical instruments and open-domain sounds. The evaluation settings are described in detail below:
   AVE~\cite{tian2018audio} contains 4143 10-second videos, including 28 diverse sound categories, such as \textit{Church Bell}, \textit{Barking}, and \textit{Frying}, among others. The AVE dataset presents greater challenges as the audio in these videos may not span the entire duration and can be noisy, including off-screen sounds (\textit{e.g.}, human speech) and background noise. 
    In addition to the AVE dataset, we also evaluate our proposed method on the widely-used MUSIC~\cite{zhao2018sound} dataset, which includes 11 musical instrument categories: accordion, acoustic guitar, cello, clarinet, erhu, flute, saxophone, trumpet, tuba, violin, and xylophone. All the videos are clean solo and the sounding instruments are visible. 
    For both datasets, we follow the same train/validation/test splits as in \cite{chen2023iquery,zhu2022visually}.

\noindent\textbf{Baselines.} To the best of our knowledge, we are the first to adopt a generative model for the audio-visual source separation task. Thus, we compare DAVIS against the following discriminative methods: \textit{NMF-MFCC}~\cite{spiertz2009source} which is an audio-only separation method; \textit{Sound of Pixels}~\cite{zhao2018sound} and \textit{Sound of Motions}~\cite{zhao2019sound} that learn ratio mask predictions with a 1-frame-based model or with motion as condition; \textit{Multisensory}~\cite{owens2018audio} that separates mixtures based on learning discriminative audio-visual representations; \textit{Minus-Plus}~\cite{xu2019recursive} that separates sounds by recursively eliminating high-energy components from the sound mixture; \textit{Cascaded Filter}~\cite{zhu2020visually} which separates sounds in a multi-stage manner; \textit{Co-Separation}~\cite{gao2019co} that takes a single visual object as the condition to perform mask regression; \textit{Cyclic Co-Learn} (CCoL)~\cite{tian2021cyclic} which jointly trains the model with sounding object visual grounding and visually-guided sound source separation tasks; \textit{AMnet}~\cite{zhu2022visually} which is a two-stage framework modeling both appearance and motion;
\textit{iQuery}~\cite{chen2023iquery} that adapts the maskformer architecture for audio-visual separation and achieves the current state-of-the-art (SOTA) results. 

\noindent\textbf{Evaluation Metrics.}
To quantitatively evaluate the audio-visual sound source separation performances, we use the standard metrics~\cite{zhao2018sound,tian2021cyclic,gao2019co}, namely: Signal-to-Distortion Ratio (SDR), Signal-to-Interference Ratio (SIR), and Signal-to-Artifact Ratio (SAR). We adopt the widely-used mir\_eval library~\cite{raffel2014mir_eval} to report the standard metrics. Note that SDR and SIR evaluate the accuracy of source separation, whereas SAR specifically measures the absence of artifacts~\cite{gao2019co}.

\begin{figure}[t]
\begin{minipage}[!t]{0.48\textwidth}
    \centering
    \scalebox{0.8}{
    \begin{tabular}{lccc|c}
        \toprule
        Block & SDR$\uparrow$ & SIR$\uparrow$ & SAR$\uparrow$  & \# Params (M)\\
        \midrule
        \{R, R, R\} & $9.03$ & $14.05$ &  $13.20$ & $51.76$ \\ 
        \{R, R, T\} & $11.78$ & $17.91$ & $15.44$ & $43.85$ \\ 
        \{R, R, TF\} & $11.50$ & $18.01$  & $15.21$ & $42.95$ \\   
        \rowcolor{mygray} \{R, TF, T\} & $11.88$ & $17.52$  & $16.12$ & $35.04$ \\   
        \bottomrule
    \end{tabular}}
    \captionof{table}{Ablation on CA block design. R, TF, and T denote ResNet, Time-Frequency, and Time Attention blocks, respectively. We highlight the setting used in this paper in \colorbox{mygray}{gray}.}
    \label{tab:ca}
\end{minipage}
\hfill
\begin{minipage}[!t]{0.45\textwidth}
    \centering
    \scalebox{0.85}{
    \begin{tabular}{lccc|c}
        \toprule
        Fusion & & SDR$\uparrow$ & SIR$\uparrow$ & SAR$\uparrow$   \\
        \midrule
        Concat & & $10.85$ &  $17.62$ & $15.52$  \\  
        FIM (Point-wise) & &$11.06$  & $17.37$  & $15.44$  \\ 
        FIM (Local) & &$11.56$  & $17.02$ & $16.28$ \\
        FIM (Global) & &$11.23$ & $17.56$  & $15.84$  \\
        \rowcolor{mygray} FIM (Local{\&}Global) & & $11.88$ & $17.52$ & $16.12$  \\ 
        \bottomrule
    \end{tabular}}
    \captionof{table}{Ablation study on Feature Interaction Module. We explore different ways of integrating audio and visual features.}
    \label{tab:fim}
\end{minipage}
\end{figure}

\begin{figure}[!ht]
\begin{minipage}[!t]{0.5\textwidth}
    \centering
    \includegraphics[width=0.99\columnwidth]{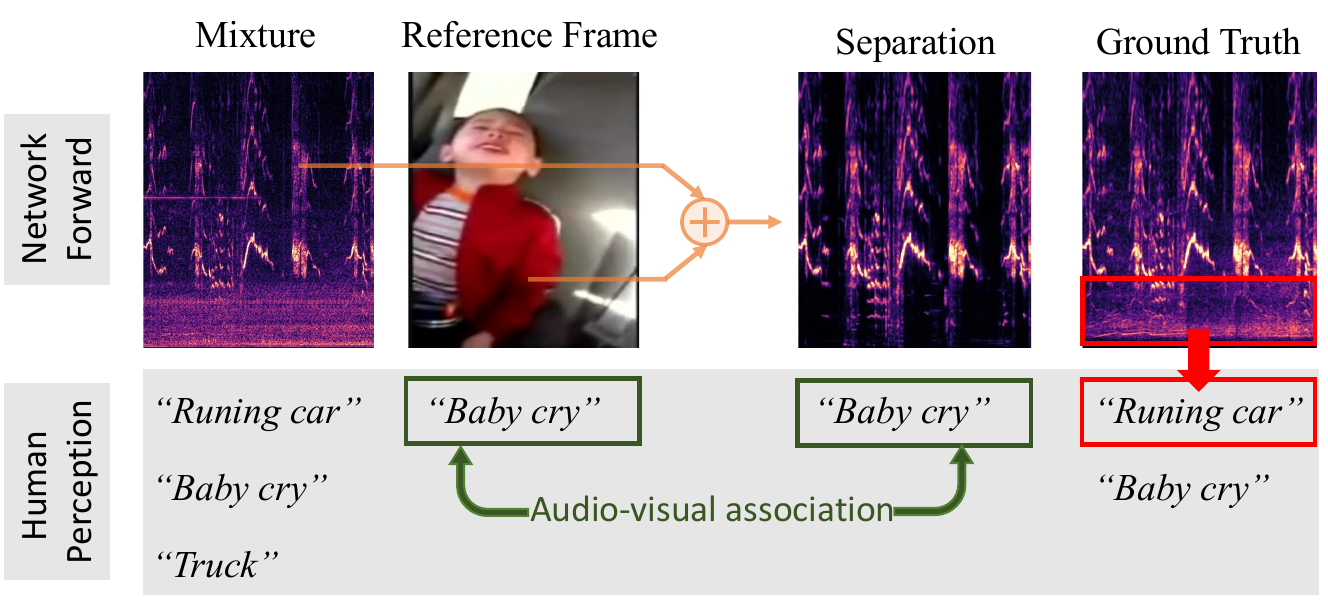}
    \captionof{figure}{A visualization example showing that our DAVIS model can capture accurate audio-visual association.}
    \label{fig:association}
\end{minipage}
\hfill
\begin{minipage}[!t]{0.45\textwidth}
    \centering
    \scalebox{1}{
    \begin{tabular}{lccc}
        \toprule
        $\delta_{silence}$ & SDR$\uparrow$ & SIR$\uparrow$ & SAR$\uparrow$   \\
        \midrule
        $0$ (baseline)  &$11.53$ &  $18.30$ & $14.74$  \\  
        \midrule
        $0.01$ & $11.15$  & $18.21$  & $14.69$  \\ 
        $0.001$ & $11.50$  & $18.26$ & $14.74$ \\
        \rowcolor{mygray} $0.002$ &$ 11.61$ & $18.36$ & $14.70$  \\ 
        \bottomrule
    \end{tabular}}
    \captionof{table}{The effect of silence mask-guided sampling strategy on the MUSIC test set.}
    \label{tab:silence}
\end{minipage}
\end{figure}

\subsection{Comparisons with State-of-the-art}
\label{subsec:comparison}

To assess the effectiveness of our method, we have compared DAVIS with state-of-the-art approaches on the AVE and MUSIC datasets. The comparison results are presented in \cref{tab:ave_music}. Our results demonstrate the benefits of using generative modeling for audio-visual separation. DAVIS achieves comparable SDR results to the strong baseline iQuery~\cite{chen2023iquery} while improving the SIR scale by \textbf{0.9 dB}. It is worth noting that the results on AVE are not as clear-cut as those on MUSIC. We believe there are two reasons behind this: firstly, iQuery uses ground truth class labels to choose a mask in both training and inference, thereby yielding stronger conditional signals. In contrast, DAVIS only uses video frames as the condition. Secondly, the original AVE audio clip often contains off-screen sounds and background noise, which the conventional metrics (\eg, SDR, SIR, and SAR) cannot accurately capture, even if the separation results have removed the background noise simultaneously. To address this issue, we provide qualitative visualization in \cref{fig:vis}, which reinforces the advantages of DAVIS. On the clean MUSIC dataset, DAVIS consistently outperforms existing methods across various evaluation metrics, surpassing the next best approach iQuery. These results clearly demonstrate DAVIS's versatility across diverse datasets with varying categories, while the subjective test in \cref{subsec:analysis} also supports our claim.

\subsection{Experimental Analysis}
\label{subsec:analysis}

We conduct ablations on the MUSIC validation set (unless specified) to examine the different components of DAVIS. \textit{For more ablations, please refer to the supplementary materials.}
\\
\textbf{Block Design.} We validate the effectiveness of our proposed CA block (shown in \cref{fig: ca}) by designing the following baselines: (a) using three consecutive ResNet blocks within the CA block, which only captures local time-frequency patterns; (b) replacing the last ResNet block with a Time Attention block; (c) replacing the last ResNet block with a Time-Frequency Attention block; and (d) replacing the last two ResNet blocks with Time-Frequency and Time attention blocks to enhance the capability of modeling long-range dependency.
The results in \cref{tab:ca} underscore the importance of learning both local and global contexts across time and frequency dimensions. Furthermore, the comparison of model sizes confirms that the improvements are not attributable to increased network capacity.

\noindent\textbf{Audio-Visual Feature Interaction.} To validate the importance of effective audio-visual association learning for this task, we conduct an ablation study on the Feature Interaction Module. Specifically, we explore different ways of feature interaction: (a) direct concatenation of visual and audio features, (b) a three-layer MLP for point-wise fusion, (c) three ResNet blocks, (d) three attention blocks, and (e) a combination of ResNet and attention blocks. The results presented in \cref{tab:fim} show that naive concatenation of audio and visual features performs significantly poorly while enabling further interaction between them improves the results. Among all the designs, our proposed module achieves the best results by considering both local and non-local contexts.

\begin{figure*}[!t]
    \centering
    \includegraphics[width=0.7\textwidth]{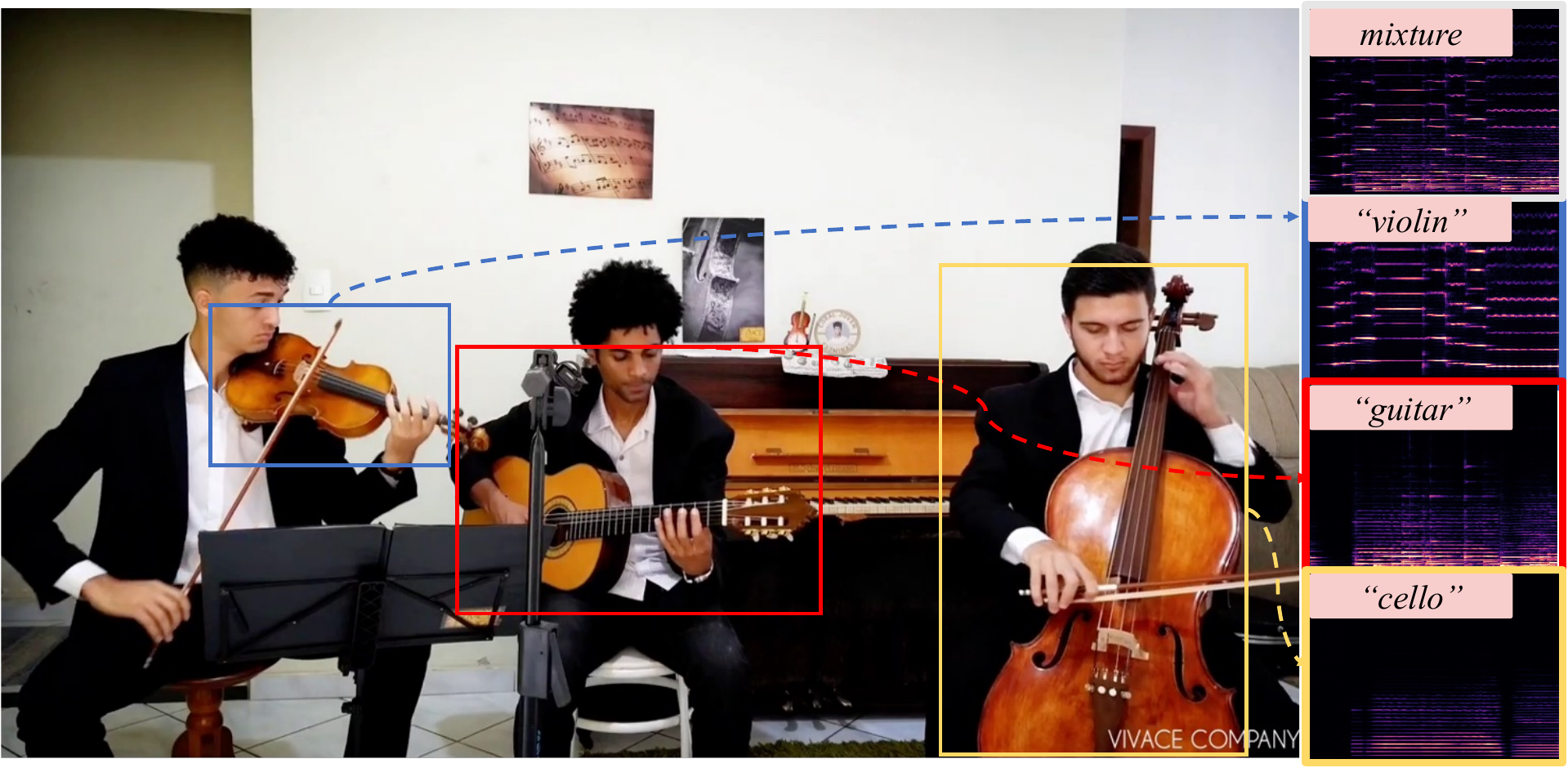}
    \caption{\textbf{Separation results on a real-world challenging three-source example}. 
    YouTube ID: \textit{R1DCTNEMibw}.
    }
    \label{fig:natural}
\end{figure*}

\noindent\textbf{Learned Audio-Visual Association.} 
To showcase the accuracy of our model's learned audio-visual associations, we mixed a ``Baby crying'' video clip with a ``Truck'' video clip from the AVE dataset.
As shown in \cref{fig:association}, the original baby video, as perceived by human listeners, also contains a running car sound, thus establishing a complicated audio-visual relationship. Our model successfully extracts the baby's crying sound while eliminating all irrelevant sounds, demonstrating DAVIS's ability to learn accurate audio-visual associations.

\begin{wrapfigure}{r}{0.2\textwidth}
    \centering
    \includegraphics[width=0.2\textwidth]{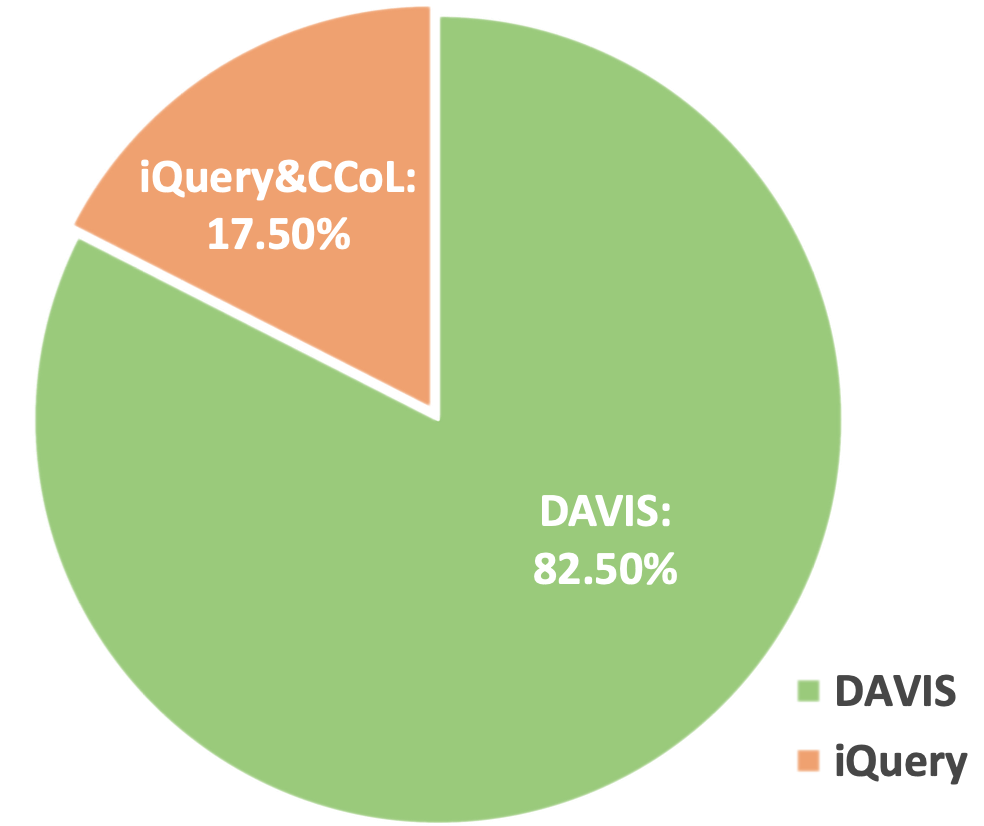}
\end{wrapfigure}

\noindent\textbf{Effects of Silence Mask-Guided Sampling.} As shown in \cref{tab:silence}, we experiment with different thresholds for the silence mask-guided sampling method, which determines the proportion of re-used information from the mixture. While a high threshold may introduce leakage from non-silent regions (\textit{e.g.}, from the second sound), we show that carefully selecting the threshold value can boost the separation performance in the post-training stage compared to the baseline.

\noindent\textbf{Qualitative Visualization on Natural Sound Mixture.}
To further demonstrate our model's effectiveness, we present a challenging real-world example of testing it on a natural sound mixture with multiple sounds (see Fig.~\ref{fig:natural}).

\noindent\textbf{Subjective test.} We conduct a subjective test of separation results of our model and strong baselines iQuery/CCoL. 11 participants are asked to answer \textit{``Which separation result is closer to the ground truth audio and better matches the frame content?''}, with GT sound and frame as a reference. DAVIS outperformed the other baselines with a \textbf{winning rate of 82.5\%} from 176 results, as shown in the figure to the right.

\subsection{Application: Zero-Shot Text-guided Separation}
\label{subsec:zeroshot}
\begin{figure}[t]
    \centering
    \includegraphics[width=0.9\textwidth]{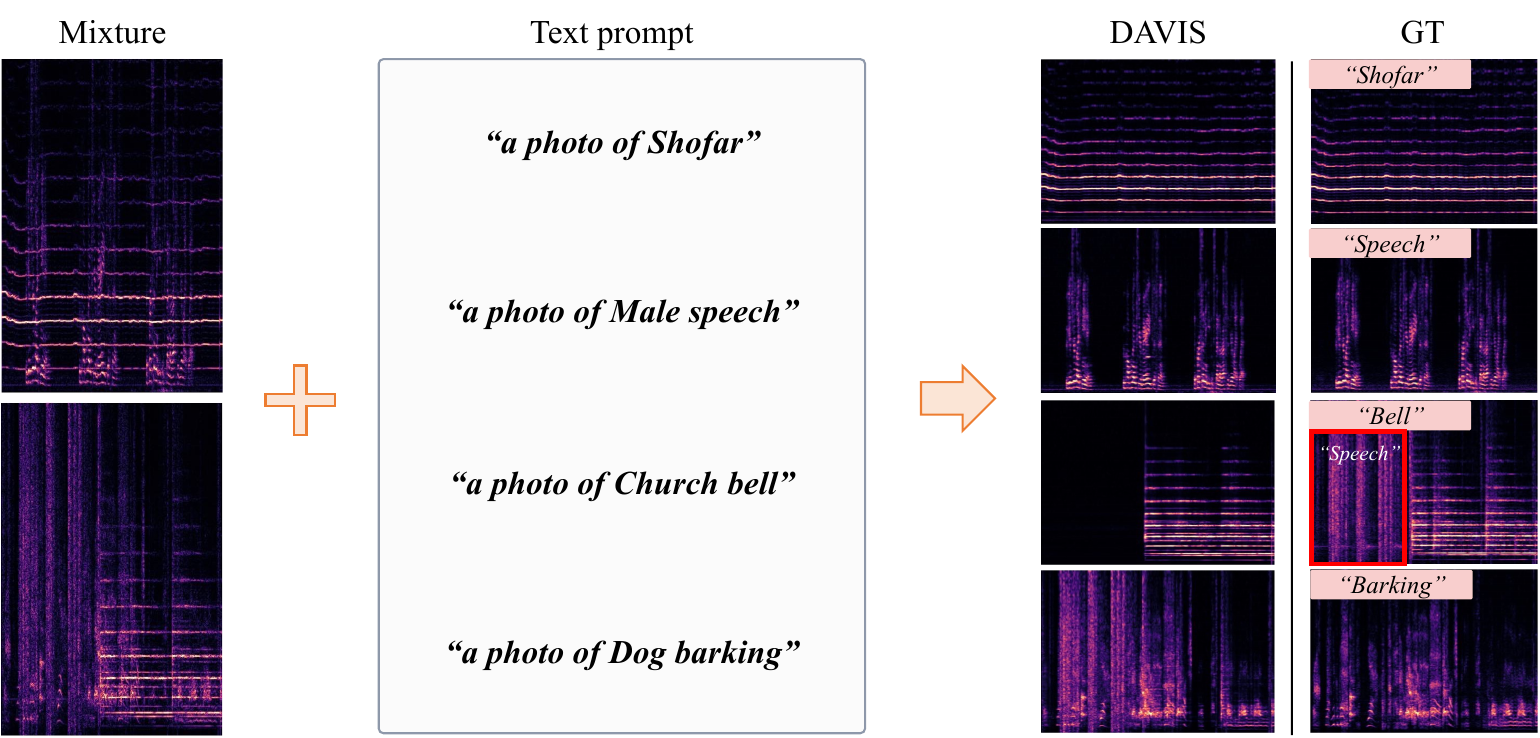}
    \caption{\textbf{Qualitative examples of zero-shot text-guided source separation.} Notably, in \textit{the third row} example, we observe the model's ability to capture precise audio-text correspondence by successfully filtering out the \textit{``speech''} sound. }
    \label{fig:textprompt}
\end{figure}

Our model trained to capture the conditional distribution $p(x|\mathbf{v})$ can be employed for zero-shot inference from $p(x|\mathbf{t})$ where $\mathbf{t}$ represents the text description corresponding to the image $\mathbf{v}$. We achieve this by leveraging the well-established shared image-text embedding space from the CLIP~\cite{radford2021learning} model. 
We qualitatively evaluate the results of utilizing replacement conditioning for separation and find it to be surprisingly effective, as shown in \cref{fig:textprompt}. 

\section{Conclusion}
In this paper, we propose DAVIS, a diffusion model-based audio-visual separation framework designed to address the problem in a generative manner.  Unlike conventional discriminative methods, DAVIS is built upon a $T$-step diffusion model, enabling the iterative synthesis of the separated magnitude spectrogram conditioned on the visual input. By proposing a specialized Separation U-Net coupled with a novel sampling strategy, we successfully apply diffusion model to this new task, yielding high-quality sound separation results.
Extensive experiments on the MUSIC and AVE datasets validate DAVIS's effectiveness in separating sounds within specific and open domains, as well as its ability to deal with diverse time-frequency structures.

\bibliographystyle{splncs04}
\bibliography{main}

\begin{thebibliography}{10}
\providecommand{\url}[1]{\texttt{#1}}
\providecommand{\urlprefix}{URL }
\providecommand{\doi}[1]{https://doi.org/#1}

\bibitem{afouras2020self}
Afouras, T., Owens, A., Chung, J.S., Zisserman, A.: Self-supervised learning of
  audio-visual objects from video. In: European Conference on Computer Vision.
  pp. 208--224. Springer (2020)

\bibitem{amit2021segdiff}
Amit, T., Shaharbany, T., Nachmani, E., Wolf, L.: Segdiff: Image segmentation
  with diffusion probabilistic models. arXiv preprint arXiv:2112.00390  (2021)

\bibitem{austin2021structured}
Austin, J., Johnson, D.D., Ho, J., Tarlow, D., van~den Berg, R.: Structured
  denoising diffusion models in discrete state-spaces. Advances in Neural
  Information Processing Systems  \textbf{34},  17981--17993 (2021)

\bibitem{avrahami2022blended}
Avrahami, O., Lischinski, D., Fried, O.: Blended diffusion for text-driven
  editing of natural images. In: Proceedings of the IEEE/CVF Conference on
  Computer Vision and Pattern Recognition. pp. 18208--18218 (2022)

\bibitem{baranchuk2021label}
Baranchuk, D., Rubachev, I., Voynov, A., Khrulkov, V., Babenko, A.:
  Label-efficient semantic segmentation with diffusion models. arXiv preprint
  arXiv:2112.03126  (2021)

\bibitem{brempong2022denoising}
Brempong, E.A., Kornblith, S., Chen, T., Parmar, N., Minderer, M., Norouzi, M.:
  Denoising pretraining for semantic segmentation. In: Proceedings of the
  IEEE/CVF Conference on Computer Vision and Pattern Recognition. pp.
  4175--4186 (2022)

\bibitem{chatterjee2022learning}
Chatterjee, M., Ahuja, N., Cherian, A.: Learning audio-visual dynamics using
  scene graphs for audio source separation. In: NeurIPS (2022)

\bibitem{chatterjee2021visual}
Chatterjee, M., Le~Roux, J., Ahuja, N., Cherian, A.: Visual scene graphs for
  audio source separation. In: Proceedings of the IEEE/CVF International
  Conference on Computer Vision. pp. 1204--1213 (2021)

\bibitem{chen2023iquery}
Chen, J., Zhang, R., Lian, D., Yang, J., Zeng, Z., Shi, J.: iquery: Instruments
  as queries for audio-visual sound separation. In: Proceedings of the IEEE/CVF
  Conference on Computer Vision and Pattern Recognition. pp. 14675--14686
  (2023)

\bibitem{chen2020wavegrad}
Chen, N., Zhang, Y., Zen, H., Weiss, R.J., Norouzi, M., Chan, W.: Wavegrad:
  Estimating gradients for waveform generation. arXiv preprint arXiv:2009.00713
   (2020)

\bibitem{chen2022diffusiondet}
Chen, S., Sun, P., Song, Y., Luo, P.: Diffusiondet: Diffusion model for object
  detection. arXiv preprint arXiv:2211.09788  (2022)

\bibitem{chen2022analog}
Chen, T., Zhang, R., Hinton, G.: Analog bits: Generating discrete data using
  diffusion models with self-conditioning. arXiv preprint arXiv:2208.04202
  (2022)

\bibitem{chen2024RAF}
Chen, Z., Gebru, I.D., Richardt, C., Kumar, A., Laney, W., Owens, A., Richard,
  A.: Real acoustic fields: An audio-visual room acoustics dataset and
  benchmark (2024)

\bibitem{chou2023av2wav}
Chou, J.C., Chien, C.M., Livescu, K.: Av2wav: Diffusion-based re-synthesis from
  continuous self-supervised features for audio-visual speech enhancement.
  arXiv preprint arXiv:2309.08030  (2023)

\bibitem{dhariwal2021diffusion}
Dhariwal, P., Nichol, A.: Diffusion models beat gans on image synthesis.
  Advances in Neural Information Processing Systems  \textbf{34},  8780--8794
  (2021)

\bibitem{dong2023clipsep}
Dong, H.W., Takahashi, N., Mitsufuji, Y., McAuley, J., Berg-Kirkpatrick, T.:
  Clipsep: Learning text-queried sound separation with noisy unlabeled videos.
  In: Proceedings of International Conference on Learning Representations
  (ICLR) (2023)

\bibitem{dumoulin2018feature}
Dumoulin, V., Perez, E., Schucher, N., Strub, F., Vries, H.d., Courville, A.,
  Bengio, Y.: Feature-wise transformations. Distill  \textbf{3}(7), ~e11 (2018)

\bibitem{elfwing2018sigmoid}
Elfwing, S., Uchibe, E., Doya, K.: Sigmoid-weighted linear units for neural
  network function approximation in reinforcement learning. Neural Networks
  \textbf{107},  3--11 (2018)

\bibitem{ephrat2018looking}
Ephrat, A., Mosseri, I., Lang, O., Dekel, T., Wilson, K., Hassidim, A.,
  Freeman, W.T., Rubinstein, M.: Looking to listen at the cocktail party: A
  speaker-independent audio-visual model for speech separation. arXiv preprint
  arXiv:1804.03619  (2018)

\bibitem{gan2020music}
Gan, C., Huang, D., Zhao, H., Tenenbaum, J.B., Torralba, A.: Music gesture for
  visual sound separation. In: Proceedings of the IEEE/CVF Conference on
  Computer Vision and Pattern Recognition. pp. 10478--10487 (2020)

\bibitem{gao2018learning}
Gao, R., Feris, R., Grauman, K.: Learning to separate object sounds by watching
  unlabeled video. In: Proceedings of the European Conference on Computer
  Vision (ECCV). pp. 35--53 (2018)

\bibitem{gao2019co}
Gao, R., Grauman, K.: Co-separating sounds of visual objects. In: Proceedings
  of the IEEE/CVF International Conference on Computer Vision. pp. 3879--3888
  (2019)

\bibitem{gong2022diffuseq}
Gong, S., Li, M., Feng, J., Wu, Z., Kong, L.: Diffuseq: Sequence to sequence
  text generation with diffusion models. arXiv preprint arXiv:2210.08933
  (2022)

\bibitem{gu2022vector}
Gu, S., Chen, D., Bao, J., Wen, F., Zhang, B., Chen, D., Yuan, L., Guo, B.:
  Vector quantized diffusion model for text-to-image synthesis. In: Proceedings
  of the IEEE/CVF Conference on Computer Vision and Pattern Recognition. pp.
  10696--10706 (2022)

\bibitem{ho2020denoising}
Ho, J., Jain, A., Abbeel, P.: Denoising diffusion probabilistic models.
  Advances in Neural Information Processing Systems  \textbf{33},  6840--6851
  (2020)

\bibitem{ho2022video}
Ho, J., Salimans, T., Gritsenko, A., Chan, W., Norouzi, M., Fleet, D.J.: Video
  diffusion models. arXiv preprint arXiv:2204.03458  (2022)

\bibitem{huang2023egocentric}
Huang, C., Tian, Y., Kumar, A., Xu, C.: Egocentric audio-visual object
  localization. arXiv preprint arXiv:2303.13471  (2023)

\bibitem{huang2022prodiff}
Huang, R., Zhao, Z., Liu, H., Liu, J., Cui, C., Ren, Y.: Prodiff: Progressive
  fast diffusion model for high-quality text-to-speech. In: Proceedings of the
  30th ACM International Conference on Multimedia. pp. 2595--2605 (2022)

\bibitem{kong2020diffwave}
Kong, Z., Ping, W., Huang, J., Zhao, K., Catanzaro, B.: Diffwave: A versatile
  diffusion model for audio synthesis. arXiv preprint arXiv:2009.09761  (2020)

\bibitem{lee2021nu}
Lee, J., Han, S.: Nu-wave: A diffusion probabilistic model for neural audio
  upsampling. arXiv preprint arXiv:2104.02321  (2021)

\bibitem{lee2023seeing}
Lee, S., Jung, C., Jang, Y., Kim, J., Chung, J.S.: Seeing through the
  conversation: Audio-visual speech separation based on diffusion model. arXiv
  preprint arXiv:2310.19581  (2023)

\bibitem{li2022diffusion}
Li, X., Thickstun, J., Gulrajani, I., Liang, P.S., Hashimoto, T.B.:
  Diffusion-lm improves controllable text generation. Advances in Neural
  Information Processing Systems  \textbf{35},  4328--4343 (2022)

\bibitem{NEURIPS2023_760dff0f}
Liang, S., Huang, C., Tian, Y., Kumar, A., Xu, C.: Av-nerf: Learning neural
  fields for real-world audio-visual scene synthesis. In: Oh, A., Naumann, T.,
  Globerson, A., Saenko, K., Hardt, M., Levine, S. (eds.) Advances in Neural
  Information Processing Systems. vol.~36, pp. 37472--37490. Curran Associates,
  Inc. (2023),
  \url{https://proceedings.neurips.cc/paper_files/paper/2023/file/760dff0f9c0e9ed4d7e22918c73351d4-Paper-Conference.pdf}

\bibitem{liang2023neural}
Liang, S., Huang, C., Tian, Y., Kumar, A., Xu, C.: Neural acoustic context
  field: Rendering realistic room impulse response with neural fields. arXiv
  preprint arXiv:2309.15977  (2023)

\bibitem{meng2021sdedit}
Meng, C., He, Y., Song, Y., Song, J., Wu, J., Zhu, J.Y., Ermon, S.: Sdedit:
  Guided image synthesis and editing with stochastic differential equations.
  arXiv preprint arXiv:2108.01073  (2021)

\bibitem{michelsanti2021overview}
Michelsanti, D., Tan, Z.H., Zhang, S.X., Xu, Y., Yu, M., Yu, D., Jensen, J.: An
  overview of deep-learning-based audio-visual speech enhancement and
  separation. IEEE/ACM Transactions on Audio, Speech, and Language Processing
  \textbf{29},  1368--1396 (2021)

\bibitem{mittal2022learning}
Mittal, H., Morgado, P., Jain, U., Gupta, A.: Learning state-aware visual
  representations from audible interactions. In: Proceedings of the European
  conference on computer vision (ECCV) (2022)

\bibitem{nichol2021glide}
Nichol, A., Dhariwal, P., Ramesh, A., Shyam, P., Mishkin, P., McGrew, B.,
  Sutskever, I., Chen, M.: Glide: Towards photorealistic image generation and
  editing with text-guided diffusion models. arXiv preprint arXiv:2112.10741
  (2021)

\bibitem{nichol2021improved}
Nichol, A.Q., Dhariwal, P.: Improved denoising diffusion probabilistic models.
  In: International Conference on Machine Learning. pp. 8162--8171. PMLR (2021)

\bibitem{owens2018audio}
Owens, A., Efros, A.A.: Audio-visual scene analysis with self-supervised
  multisensory features. In: Proceedings of the European Conference on Computer
  Vision (ECCV). pp. 631--648 (2018)

\bibitem{popov2021grad}
Popov, V., Vovk, I., Gogoryan, V., Sadekova, T., Kudinov, M.: Grad-tts: A
  diffusion probabilistic model for text-to-speech. In: International
  Conference on Machine Learning. pp. 8599--8608. PMLR (2021)

\bibitem{qian2020multiple}
Qian, R., Hu, D., Dinkel, H., Wu, M., Xu, N., Lin, W.: Multiple sound sources
  localization from coarse to fine. In: European Conference on Computer Vision.
  pp. 292--308. Springer (2020)

\bibitem{radford2021learning}
Radford, A., Kim, J.W., Hallacy, C., Ramesh, A., Goh, G., Agarwal, S., Sastry,
  G., Askell, A., Mishkin, P., Clark, J., et~al.: Learning transferable visual
  models from natural language supervision. In: International conference on
  machine learning. pp. 8748--8763. PMLR (2021)

\bibitem{raffel2014mir_eval}
Raffel, C., McFee, B., Humphrey, E.J., Salamon, J., Nieto, O., Liang, D.,
  Ellis, D.P., Raffel, C.C.: Mir\_eval: A transparent implementation of common
  mir metrics. In: ISMIR. pp. 367--372 (2014)

\bibitem{ramesh2022hierarchical}
Ramesh, A., Dhariwal, P., Nichol, A., Chu, C., Chen, M.: Hierarchical
  text-conditional image generation with clip latents. arXiv preprint
  arXiv:2204.06125  (2022)

\bibitem{rombach2022high}
Rombach, R., Blattmann, A., Lorenz, D., Esser, P., Ommer, B.: High-resolution
  image synthesis with latent diffusion models. In: Proceedings of the IEEE/CVF
  conference on computer vision and pattern recognition. pp. 10684--10695
  (2022)

\bibitem{ronneberger2015u}
Ronneberger, O., Fischer, P., Brox, T.: U-net: Convolutional networks for
  biomedical image segmentation. In: Medical Image Computing and
  Computer-Assisted Intervention--MICCAI 2015: 18th International Conference,
  Munich, Germany, October 5-9, 2015, Proceedings, Part III 18. pp. 234--241.
  Springer (2015)

\bibitem{ruan2022mm}
Ruan, L., Ma, Y., Yang, H., He, H., Liu, B., Fu, J., Yuan, N.J., Jin, Q., Guo,
  B.: Mm-diffusion: Learning multi-modal diffusion models for joint audio and
  video generation. arXiv preprint arXiv:2212.09478  (2022)

\bibitem{ruiz2022dreambooth}
Ruiz, N., Li, Y., Jampani, V., Pritch, Y., Rubinstein, M., Aberman, K.:
  Dreambooth: Fine tuning text-to-image diffusion models for subject-driven
  generation. arXiv preprint arXiv:2208.12242  (2022)

\bibitem{saharia2022photorealistic}
Saharia, C., Chan, W., Saxena, S., Li, L., Whang, J., Denton, E.L.,
  Ghasemipour, K., Gontijo~Lopes, R., Karagol~Ayan, B., Salimans, T., et~al.:
  Photorealistic text-to-image diffusion models with deep language
  understanding. Advances in Neural Information Processing Systems
  \textbf{35},  36479--36494 (2022)

\bibitem{scheibler2023diffusion}
Scheibler, R., Ji, Y., Chung, S.W., Byun, J., Choe, S., Choi, M.S.:
  Diffusion-based generative speech source separation. In: ICASSP 2023-2023
  IEEE International Conference on Acoustics, Speech and Signal Processing
  (ICASSP). pp.~1--5. IEEE (2023)

\bibitem{shen2021efficient}
Shen, Z., Zhang, M., Zhao, H., Yi, S., Li, H.: Efficient attention: Attention
  with linear complexities. In: Proceedings of the IEEE/CVF winter conference
  on applications of computer vision. pp. 3531--3539 (2021)

\bibitem{singer2022make}
Singer, U., Polyak, A., Hayes, T., Yin, X., An, J., Zhang, S., Hu, Q., Yang,
  H., Ashual, O., Gafni, O., et~al.: Make-a-video: Text-to-video generation
  without text-video data. arXiv preprint arXiv:2209.14792  (2022)

\bibitem{smaragdis2003non}
Smaragdis, P., Brown, J.C.: Non-negative matrix factorization for polyphonic
  music transcription. In: 2003 IEEE Workshop on Applications of Signal
  Processing to Audio and Acoustics (IEEE Cat. No. 03TH8684). pp. 177--180.
  IEEE (2003)

\bibitem{song2020denoising}
Song, J., Meng, C., Ermon, S.: Denoising diffusion implicit models. arXiv
  preprint arXiv:2010.02502  (2020)

\bibitem{song2019generative}
Song, Y., Ermon, S.: Generative modeling by estimating gradients of the data
  distribution. Advances in neural information processing systems  \textbf{32}
  (2019)

\bibitem{song2020score}
Song, Y., Sohl-Dickstein, J., Kingma, D.P., Kumar, A., Ermon, S., Poole, B.:
  Score-based generative modeling through stochastic differential equations.
  arXiv preprint arXiv:2011.13456  (2020)

\bibitem{spiertz2009source}
Spiertz, M., Gnann, V.: Source-filter based clustering for monaural blind
  source separation. In: Proceedings of the 12th International Conference on
  Digital Audio Effects. vol.~4, p.~6 (2009)

\bibitem{tan2023language}
Tan, R., Ray, A., Burns, A., Plummer, B.A., Salamon, J., Nieto, O., Russell,
  B., Saenko, K.: Language-guided audio-visual source separation via trimodal
  consistency. arXiv preprint arXiv:2303.16342  (2023)

\bibitem{tian2021cyclic}
Tian, Y., Hu, D., Xu, C.: Cyclic co-learning of sounding object visual
  grounding and sound separation. In: Proceedings of the IEEE/CVF Conference on
  Computer Vision and Pattern Recognition. pp. 2745--2754 (2021)

\bibitem{tian2018audio}
Tian, Y., Shi, J., Li, B., Duan, Z., Xu, C.: Audio-visual event localization in
  unconstrained videos. In: Proceedings of the European Conference on Computer
  Vision (ECCV). pp. 247--263 (2018)

\bibitem{tzinis2020into}
Tzinis, E., Wisdom, S., Jansen, A., Hershey, S., Remez, T., Ellis, D.P.,
  Hershey, J.R.: Into the wild with audioscope: Unsupervised audio-visual
  separation of on-screen sounds. arXiv preprint arXiv:2011.01143  (2020)

\bibitem{tzinis2022audioscopev2}
Tzinis, E., Wisdom, S., Remez, T., Hershey, J.R.: Audioscopev2: Audio-visual
  attention architectures for calibrated open-domain on-screen sound
  separation. In: Computer Vision--ECCV 2022: 17th European Conference, Tel
  Aviv, Israel, October 23--27, 2022, Proceedings, Part XXXVII. pp. 368--385.
  Springer (2022)

\bibitem{vaswani2017attention}
Vaswani, A., Shazeer, N., Parmar, N., Uszkoreit, J., Jones, L., Gomez, A.N.,
  Kaiser, {\L}., Polosukhin, I.: Attention is all you need. Advances in neural
  information processing systems  \textbf{30} (2017)

\bibitem{virtanen2007monaural}
Virtanen, T.: Monaural sound source separation by nonnegative matrix
  factorization with temporal continuity and sparseness criteria. IEEE
  transactions on audio, speech, and language processing  \textbf{15}(3),
  1066--1074 (2007)

\bibitem{wang2023tf}
Wang, Z.Q., Cornell, S., Choi, S., Lee, Y., Kim, B.Y., Watanabe, S.:
  Tf-gridnet: Making time-frequency domain models great again for monaural
  speaker separation. In: ICASSP 2023-2023 IEEE International Conference on
  Acoustics, Speech and Signal Processing (ICASSP). pp.~1--5. IEEE (2023)

\bibitem{wu2018group}
Wu, Y., He, K.: Group normalization. In: Proceedings of the European conference
  on computer vision (ECCV). pp. 3--19 (2018)

\bibitem{xu2019recursive}
Xu, X., Dai, B., Lin, D.: Recursive visual sound separation using minus-plus
  net. In: Proceedings of the IEEE/CVF International Conference on Computer
  Vision. pp. 882--891 (2019)

\bibitem{zhao2019sound}
Zhao, H., Gan, C., Ma, W.C., Torralba, A.: The sound of motions. In:
  Proceedings of the IEEE/CVF International Conference on Computer Vision. pp.
  1735--1744 (2019)

\bibitem{zhao2018sound}
Zhao, H., Gan, C., Rouditchenko, A., Vondrick, C., McDermott, J., Torralba, A.:
  The sound of pixels. In: Proceedings of the European conference on computer
  vision (ECCV). pp. 570--586 (2018)

\bibitem{zhu2020visually}
Zhu, L., Rahtu, E.: Visually guided sound source separation using cascaded
  opponent filter network. In: Proceedings of the Asian Conference on Computer
  Vision (2020)

\bibitem{zhu2022visually}
Zhu, L., Rahtu, E.: Visually guided sound source separation and localization
  using self-supervised motion representations. In: Proceedings of the IEEE/CVF
  Winter Conference on Applications of Computer Vision. pp. 1289--1299 (2022)

\end{thebibliography}
\end{document}